\documentclass[sigconf]{acmart}
\usepackage{algorithm}
\usepackage{algorithmic}
\usepackage{babel,blindtext}
\AtBeginDocument{%
  \providecommand\BibTeX{{%
    \normalfont B\kern-0.5em{\scshape i\kern-0.25em b}\kern-0.8em\TeX}}}

\setcopyright{acmcopyright}
\copyrightyear{2021}
\acmYear{2021}
\acmDOI{10.1145/1122445.1122456}

\acmConference[Ubicomp21]{ UbiComp '21: ACM Conference on Pervasive and Ubiquitous Computing (UbiComp)}{Sep 21--26, 2021}{Virtual}
\acmBooktitle{Conference on Pervasive and Ubiquitous Computing}
\begin{document}
\sloppy
\title{Package Delivery Using Autonomous Drones in Skyways}

\author{Woojin Lee}
\affiliation{%
  \institution{The University of Sydney}
  \country{Australia}}
\email{wlee2926@uni.sydney.edu.au}

\author{Balsam Alkouz}
\affiliation{%
  \institution{The University of Sydney}
  \country{Australia}}
\email{balsam.alkouz@sydney.edu.au}

\author{Babar Shazaad}
\affiliation{%
\institution{The University of Sydney}
  \country{Australia}
}
\email{babar.shahzaad@sydney.edu.au}

\author{Athman Bouguettaya}
\affiliation{%
 \institution{The University of Sydney}
 \country{Australia}}
\email{athman.bouguettaya@sydney.edu.au}


\begin{abstract}
Current drone delivery systems mostly focus on point-to-point package delivery. We present a multi-stop drone service system to deliver packages anywhere anytime within a specified geographic area. We define a skyway network which takes into account flying regulations, including restricted areas and no-fly zones. The skyway nodes typically represent building rooftops which may act as both recharging stations and delivery destinations.
% We take into consideration no-flight zone that could be specified by governments.
A heuristic-based A* algorithm is used to compute an optimal path from source to destination taking into account a number of constraints, including delivery time, availability of recharging stations, etc. We deploy our drone delivery system in an indoor testbed environment using a 3D model of Sydney CBD. We describe a graphical user interface to monitor the real-time package delivery in the skyway network.

% A drone utilizes land marks to locate and position itself in the indoor testbed environment built of Sydney's CBD. Finally, a user interface is designed to track the drone during its delivery mission.

% Current drone delivery systems assume point to point deliveries only. We present a multi-stop drone delivery platform in a smart city. We leverage building rooftops and propose a skyway network to connect the rooftops. We demonstrate full delivery processes from sources to destinations using the network. We take into consideration the constraints surrounding drone deliveries. We leverage a modified A* algorithm to compose the path and navigate the drone.

\end{abstract}





\maketitle

\section{Introduction}

Drones are unmanned aircraft that operate with various degrees of autonomy. They have seen phenomenal growth and uptake which has come as a result of the drop in their prices and increased sophistication \cite{chmaj2015distributed}. The wide availability of drones opens opportunities for a wide number of applications, such as emergency response, surveillance, and package delivery \cite{shakhatreh2019unmanned}.
% The focus of this demo is on operating drones for the delivery of goods.
Using drones in \emph{delivery services} is a fast-growing industry that gained a lot of commercial attention from companies such as Amazon, DHL, and Google \cite{bamburry2015drones}. COVID-19 pandemic has highlighted the need for such technologies as drones to ensure that economic activities continue to thrive. Drones provide a safer, socially distanced, contactless, and more resilient alternative to deliver goods in cities and regional areas \cite{KUMAR20211}. 

% The key \emph{beneficiaries} of drone-based delivery services include end-users (consumers), transport companies, and supplier of goods (e.g., medical suppliers, retailers, etc.).

\begin{figure}[t]
\centering
\includegraphics[width=0.6\linewidth]{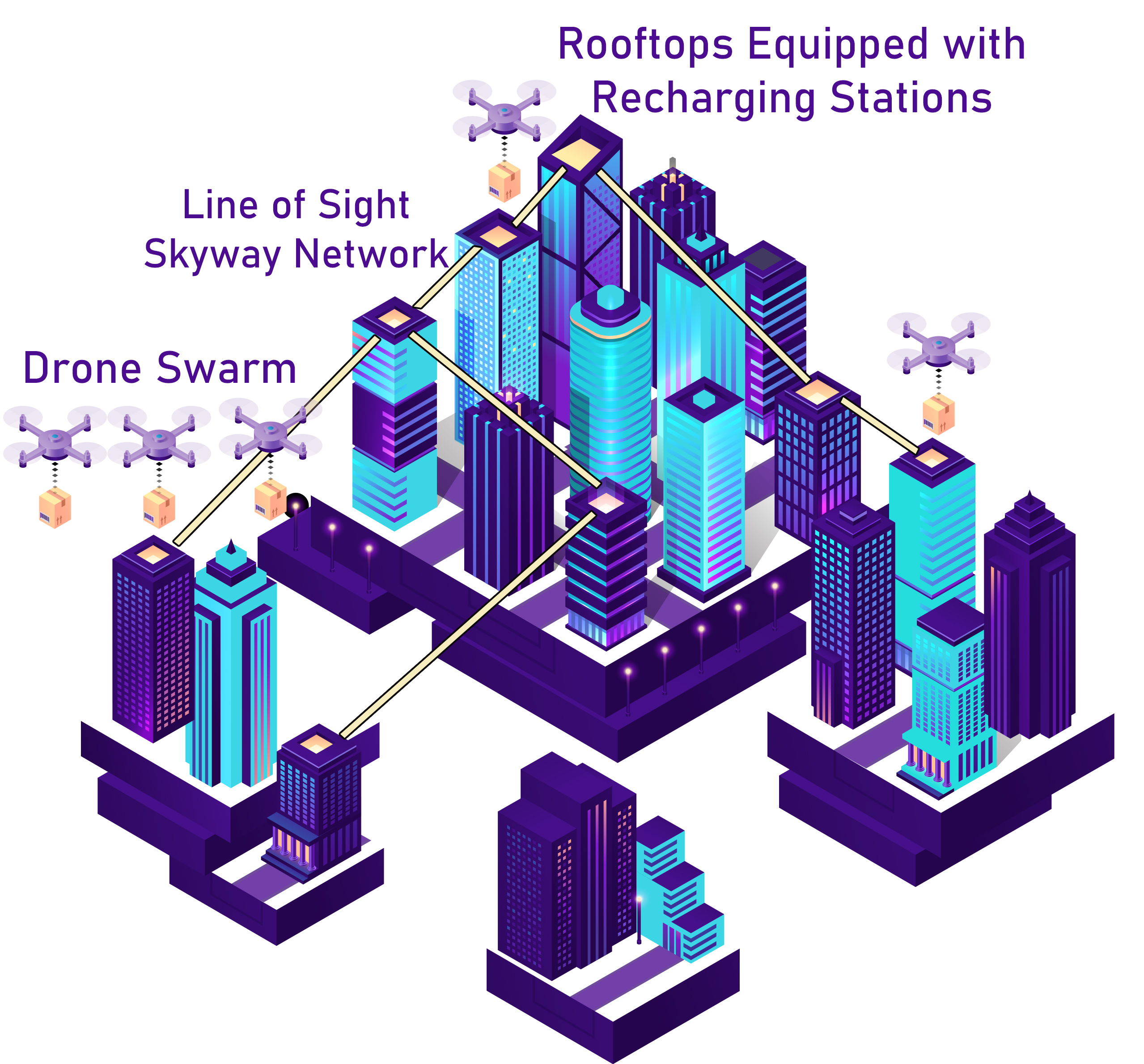}
\vspace{-0.4 cm}
\caption{Drone Delivery in a Skyway Network}
\label{skyway}
\vspace{-0.7 cm}
\end{figure}

This demo focuses on developing a novel system to effectively provision \emph{drone-based delivery services} in a skyway network. A skyway network enables the safe and scalable deployment of drone-based delivery solutions in shared airspace \cite{shahzaad2019composing}. Fig. \ref{skyway} depicts the drone delivery in a skyway network where nodes are the building rooftops.
% that links a set of \emph{line segments} between any two nodes within line of sight (LoS). Each node is a fixed landing pad on a rooftop of a building.
We use \emph{existing infrastructure of a city} where each building rooftop may be easily and cheaply fitted with a recharging pad.

\begin{figure*}[!htb]

\minipage{0.32\textwidth}
  \includegraphics[width=\linewidth]{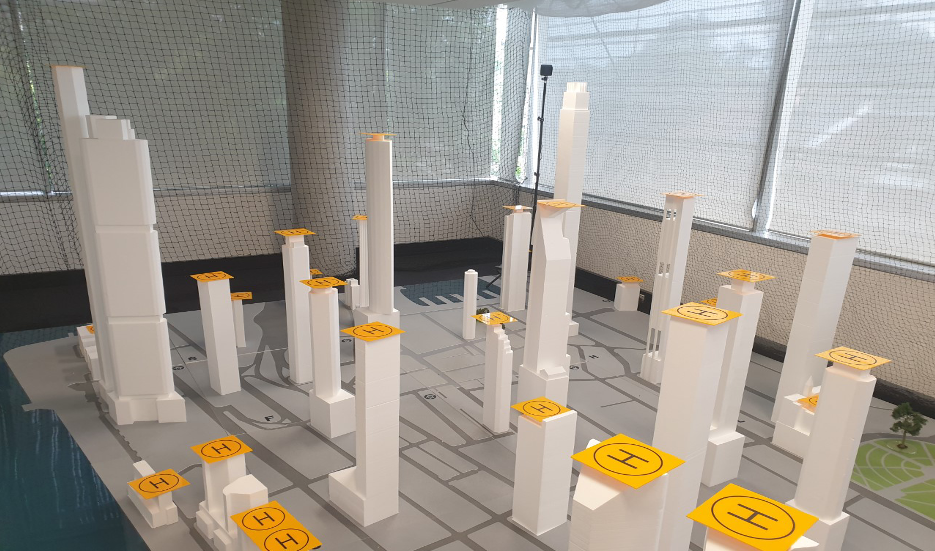}
  \caption{3D Model of Sydney CBD}\label{3d}
\endminipage\hfill
\minipage{0.32\textwidth}%
  \includegraphics[width=0.8\linewidth]{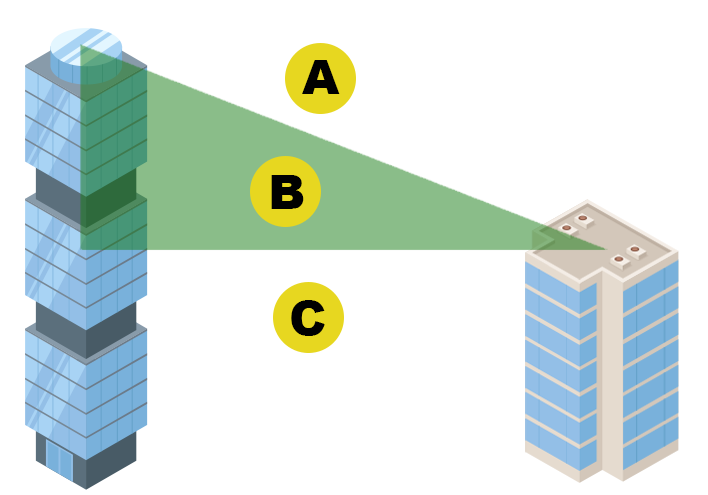}
  \caption{Line of Sight Intersection}\label{los}
\endminipage\hfill
\minipage{0.32\textwidth}
  \includegraphics[width=\linewidth]{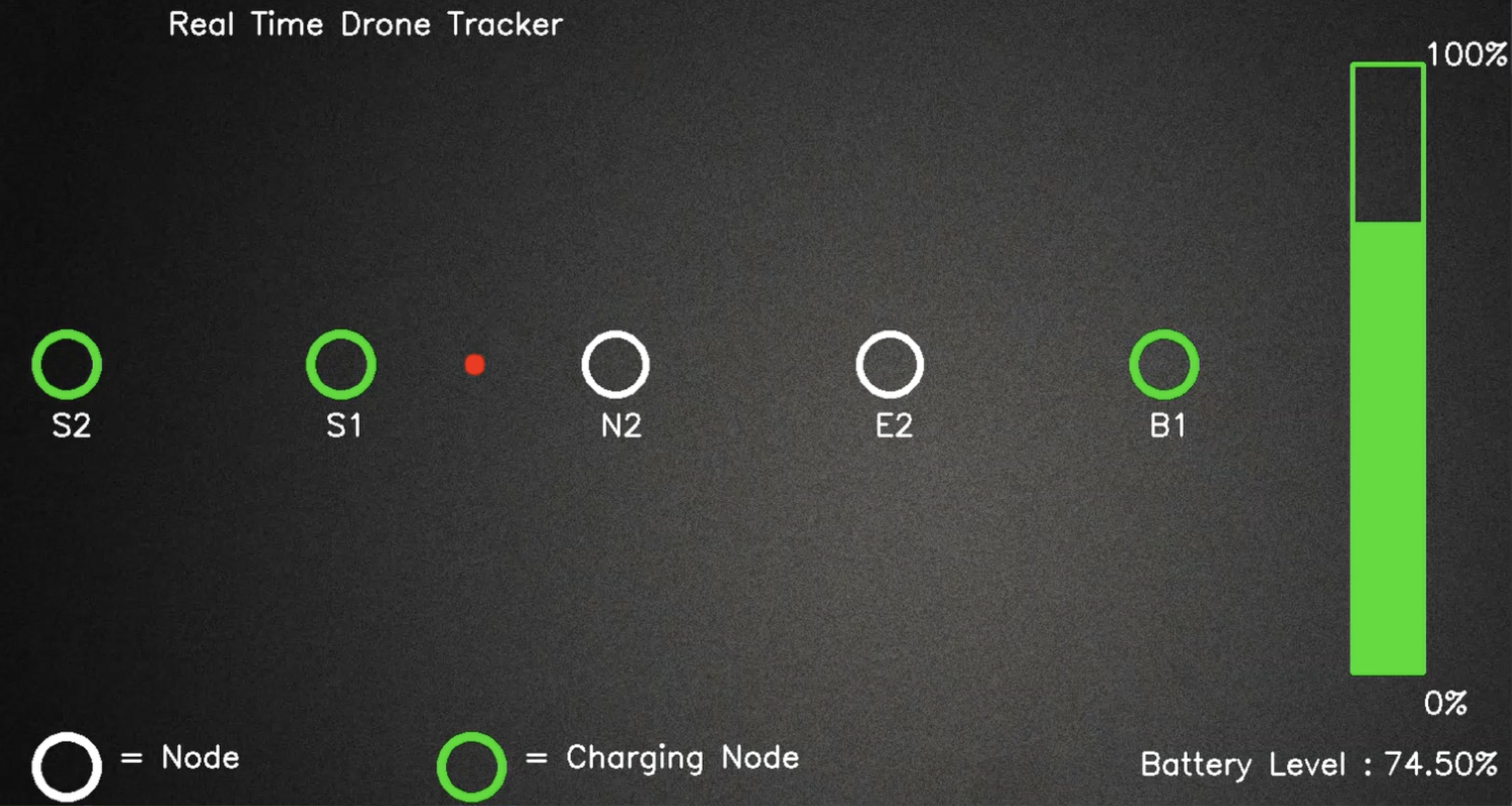}
  \caption{GUI Screenshot}\label{GUI}
\endminipage
\vspace{-0.4 cm}
\end{figure*}

% \section{Related Work}
% The common denominator for the literature focuses on drone delivery services operating along a single segment (i.e., route) from a source to a destination \cite{dorling2016vehicle} \cite{sundar2013algorithms}. The selection and composition of efficient delivery services in a multi-constrained skyway network are not focused in the existing works. An efficient and cost-effective drone service framework is required which incorporates the real-world aspects of drone delivery services in terms of intrinsic and extrinsic constraints and execution of services along multi-segment paths in a skyway network. This demo proposes to augment and extend the existing works in a multi-route skyway network.
 
\section{Demonstration System}
\subsection{Environment and Hardware Components}
Having an outdoor test environment is fraught with risks because of government restrictions when using drones in built-up areas \cite{jones2017international}. Therefore, we setup a 3D model of Sydney CBD as an indoor drone testbed at the University of Sydney (Fig. \ref{3d}). We build a skyway network which contains virtual barriers to prevent drones’ access to no-fly zones and restricted areas. A \emph{skyway network} is defined as a set of connected nodes representing take-off and landing stations. Each node may concurrently act as a recharging station. We use the LoS-based paths computed in subsection \ref{lineofsight} to perform drone-based deliveries in the skyway network from a given source to a destination. We use a \emph{DJI Tello EDU} drone that is safe and well suited to fly in an indoor testbed for its small size and robust nature.

% We print 3D extended legs for the drones that weight 3 grams. The purpose of the extended legs is to allow a space to attach the payload to the drone when it is on land. 
% We attach a small package that weights 5 grams to the drone to demonstrate the delivery process.

% \begin{figure}[ht]
% \centering
% \includegraphics[width=0.9\linewidth]{Figures/3D model.png}
% \caption{3D Model of Sydney CBD}
% \label{3d}
% \end{figure}

% \begin{table}[]
% \caption{Drone Specifications}
% \label{tab:my-table}
% \begin{tabular}{l|l}
% \hline
% Variable            & Value         \\ \hline
% Weight              & 80 g          \\
% Dimensions          & 98×92.5×41 mm \\
% Propeller           & 3 inches      \\
% Max Flight Distance & 100m          \\
% Max Speed           & 8m/s          \\
% Max Flight Time     & 13min         \\
% Max Flight Height   & 30m           \\
% Battery             & 1.1Ah/3.8V    \\ \hline
% \end{tabular}
% \end{table}
\vspace{-0.2 cm}
\subsection{Software Components}
The software consists of four components as follows. The first component computes the LoS paths among all building rooftops to construct a skyway network. Then, the second component computes an optimal path from a given source to a destination using a heuristic-based A* algorithm. The third component is used to navigate the drone following the LoS paths in the skyway network. Finally, the fourth component presents a graphical user interface to monitor the drone position and battery status during the delivery operation.

%The battery consumption rate of the loaded Tello drone is 1.5\% per 20 seconds of the battery in-flight. However, for the sake of the demo, and since the indoor testbed is small in size, we assume the battery consumption rate is 20\% per second. This is essential to illustrate the multi-stop path composition in a skyway network.
\vspace{-0.2 cm}
\subsubsection{\textbf{Skyway Network}}
\label{lineofsight}
We construct a skyway network using the (x, y, z) coordinates of the building rooftops. The skyway network takes into account the drone flying regulations, including flying within LoS and avoiding no-fly zones. The LoS is defined as a \emph{direct line segment} between any two building rooftops having no barrier in between. We develop a LoS algorithm to compute the LoS paths between building rooftops considering no-fly zones. The no-fly zones are usually defined by aviation authorities for security and public safety, e.g., CASA in Australia\footnote{https://www.casa.gov.au/drones/drone-rules/flying-near-emergencies-and-public-spaces}. We specify seven no-fly zones where drones are not permitted to fly (as shown in Fig. \ref{skyway}).

% Given the buildings coordinates, the purpose of this module is to create a LoS skyway network connecting the buildings rooftops. An LoS is a segment without intersecting buildings in between. We develop an LoS algorithm that computes the LoS paths between the buildings considering no-flight zones. A no flight zone is an area that drones are not allowed to fly in due to security or public safety reasons\footnote{https://www.casa.gov.au/drones/drone-rules/flying-near-emergencies-and-public-spaces}. We specify 7 no-flight zones as shown in Fig. \ref{skyway}

% \begin{figure}[ht]
% \centering
% \includegraphics[width=0.7\linewidth]{Figures/line of sight copy.png}
% \caption{Line of Sight Intersection}
% \label{los}
% \end{figure}

The LoS algorithm detects barriers (i.e., intersecting buildings) between any two nodes (i.e., building rooftops) to form a skyway path between them.
% The z coordinate is considered as the highest point of the building where the drone would typically land or hover to recharge.
We first construct a rectangular polygon to represent a path between any two nodes taking into account the drone width. Then, we discover and mark all nodes intersecting with the rectangular polygon considering only the xy-plane. The initial two steps are used to reduce the search space and computation cost by selecting only feasible nodes. We then construct a right angle triangle between any two buildings using their heights which are represented by z-axis (as shown in Fig. \ref{los}). The yellow circled nodes indicate nodes which intersect with the rectangular polygon considering their z-axis. The nodes above than the right angle triangle obstruct the LoS path (e.g., node A in this scenario). While the nodes below or within the right angle triangle do not obstruct the LoS path (e.g., node B and node C). We define a LoS-based straight path between any two nodes if no node obstructs the path. This process continues until all possible LoS paths are computed and connected to form the skyway network.

% The detection process for each skyway path is accomplished in several steps.

% At a node, there are several steps involved to determine if a node is directly reachable from the current node. First, for the given drone width, construct a rectangular polygon that represents the path the drone will take between the source and the destination. Second, find all nodes in the area that intersect with this polygon and mark them. This step only considers the xy-plane of the node. The goal of the first two steps is to reduce the computation cost and only consider feasible nodes. Next, as shown in Fig.\ref{los}, a right angle triangle polygon is constructed using the source and destination heights (z coordinates). Consider the yellow circles as the z coordinates of the nodes that are identified to be intersecting the first polygon. Then, if the point is above the green polygon, it will obstruct the path (point A). If the point is below the green polygon, or is intersecting the green polygon, it will not obstruct the path (points B and C). After the above procedure, if no nodes are found to be obstructing the path, the source and the destination can be connected, and the path is added as a segment.

\begin{figure}[ht]
\centering
\includegraphics[width=0.9\linewidth]{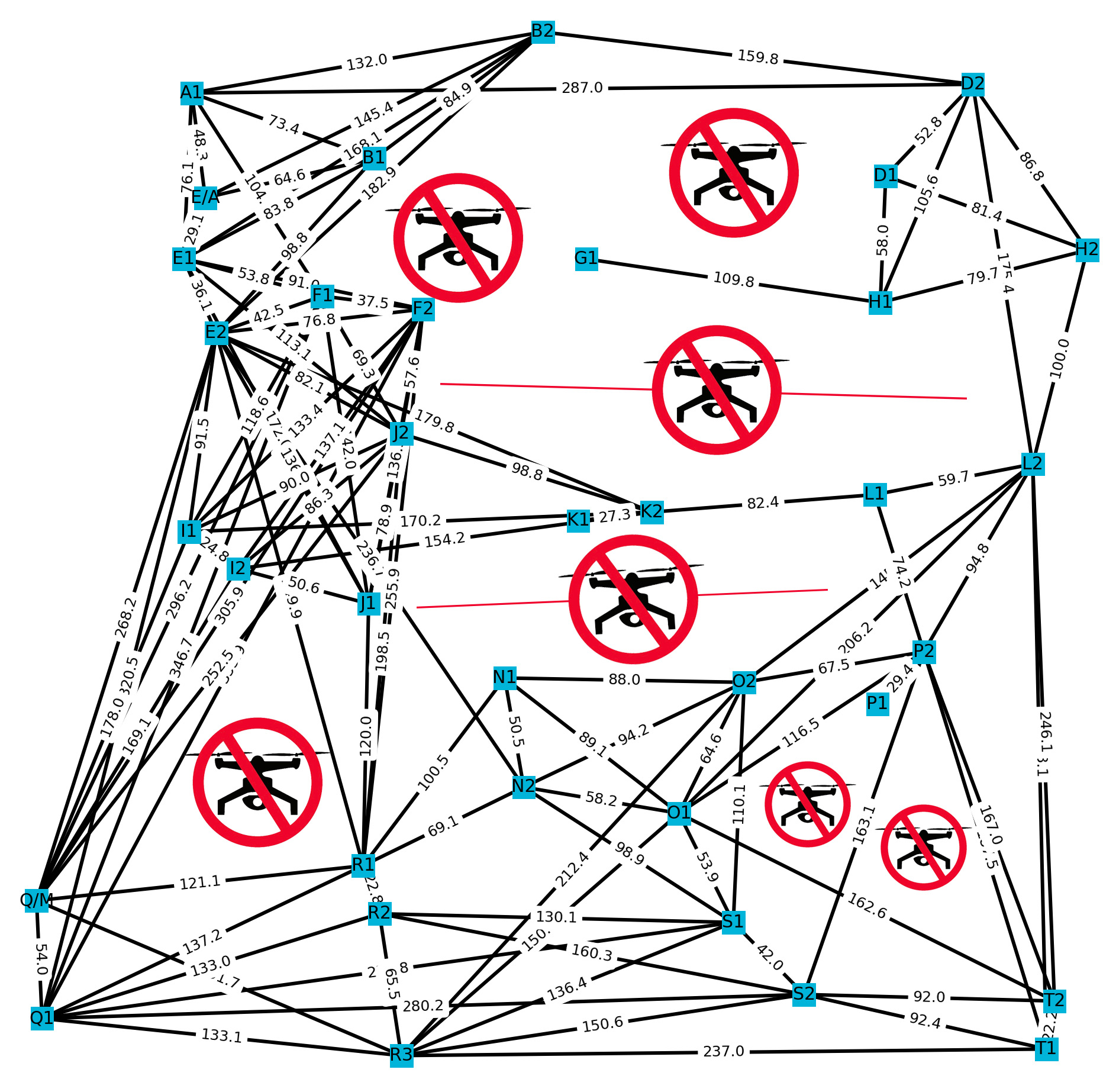}
\vspace{-0.4 cm}
\caption{A Skyway Network with No-fly Zones}
\label{motivitaing}
\vspace{-0.6 cm}
\end{figure}

\vspace{-0.2 cm}
\subsubsection{\textbf{LoS Path Composition}}
\label{pathcomposition}

We compute a temporal optimal path from the source to the destination using a heuristic-based A* algorithm \cite{alkouz2020swarm}. In this context, temporal optimal path refers to a path that leads to the destination \emph{faster}. The drone may need recharging if it cannot reach the destination with its current battery capacity. In such a case, we compose a set of LoS paths leading the package to the destination including intermediate recharging stations.

% In a \emph{LoS path composition}, the drone at the source node traverses the network to the destination using a modified A* Algorithm \cite{alkouz2020swarm}. %While the drone is not at the destination node the drone computes the potential to reach the destination from its current node without stopping to recharge. 
%We compute the potential based on the payload which affects the battery level and the total distance a drone can travel. The distance is Dijkstra's shortest distance between the two nodes. 
%If the destination is reachable then the drone traverses the nodes until it reaches the destination. If the destination is unreachable then the drone finds the nearest reachable neighbor node from the current node with the minimum travel time and recharging time. The drone then again tries to find if the destination is reachable directly until the drone is at the destination node. 
% The output of this module is the composed path from the source to the destination consisting of a set of intermediate nodes including the recharging nodes.

\vspace{-0.2 cm}
\subsubsection{\textbf{Flight Navigation}}
We implement our drone delivery system and navigate an autonomous flight using the DJI Tello EDU drone. The Tello EDU comes with a set of mission pads to make precise movements. Each mission pad contains a unique ID pattern for its identification. At the source, initial position coordinates of the drone are set to (0, 0, 0). Then, the drone detects the next node position in the path computed using the optimal path composition (subsection \ref{pathcomposition}). If the drone is not facing towards the next node, it rotates in the direction of next node and travels over the LoS path. This navigation process continues until all the nodes in the optimal path are visited and package is delivered to the destination.

% The mission pads are used to position the drone in the environment. The drone is capable of reading the mission pads. Each mission pad has unique 3D coordinates, so the drone can distinguish between pads. The drone at the source sets its coordinates to (0,0,0) and computes the path to the next node in the composed path using the path composition module (subsection \ref{pathcomposition}). First, the drone at node A rotates to face node B using the nodes coordinates. Next, the drone travels in a straight line on the LoS segment. At every node, the drone centers itself with the aid of the mission pad before moving to the next node.
\vspace{-0.2 cm}
\subsubsection{\textbf{Graphical User Interface (GUI)}}
We create a simple GUI to monitor the drone position during the delivery operation. The GUI prompts the user to input the source and the destination. Then, the GUI presents an optimal path to the destination along with drone's current position and battery status (Fig. \ref{GUI}).
% It is paramount to observe the drone's performance when flying in cities.
%Fig. \ref{GUI} presents the GUI designed for our drone delivery system.
%Using the composition algorithm in subsection \ref{pathcomposition}, the nodes in the composed path are visualized as circles and the drone movement is tracked. 
%Once the flight is initiated, the interface shows a filled circle representing the drone and the battery status of the drone. As the drone traverses the nodes to the destination, the drone icon moves along the visualized drones and the battery consumption and refill at each node. 
% On a larger scale, such GUI is essential to keep track of the drones position as its flying in a city. Fig.\ref{GUI} shows a screenshot of the GUI during the flight.

% \section{Conclusion}
% In this work, we presented a prototype that enables drone delivery in a congested city setting. We constructed a line of sight skyway network that enables a safe drone delivery process. We took into consideration no-flight zone that could be specified by governments. A modified A* composition algorithm is used to compute the optimal path from source to destination in terms of delivery time. A drone utilizes land marks to locate and position itself in the indoor test bed environment built of Sydney's CBD. Finally, a user interface is designed to track the drone during its delivery mission. For the future, extrinsic constraints, such as wind, could be simulated using fans and obstacles avoidance algorithms could be incorporated.

\section*{Acknowledgment}
This research was partly made possible by DP160103595 and LE180100158 grants from the Australian Research Council. The statements made herein are solely the responsibility of the authors.

\bibliographystyle{ACM-Reference-Format}
\bibliography{sample-base}


\begin{thebibliography}{7}

%%% the Web address.

\ifx \showCODEN    \undefined \def \showCODEN     #1{\unskip}     \fi
\ifx \showDOI      \undefined \def \showDOI       #1{#1}\fi
\ifx \showISBNx    \undefined \def \showISBNx     #1{\unskip}     \fi
\ifx \showISBNxiii \undefined \def \showISBNxiii  #1{\unskip}     \fi
\ifx \showISSN     \undefined \def \showISSN      #1{\unskip}     \fi
\ifx \showLCCN     \undefined \def \showLCCN      #1{\unskip}     \fi
\ifx \shownote     \undefined \def \shownote      #1{#1}          \fi
\ifx \showarticletitle \undefined \def \showarticletitle #1{#1}   \fi
\ifx \showURL      \undefined \def \showURL       {\relax}        \fi
% invisible to TeX
\providecommand\bibfield[2]{#2}
\providecommand\bibinfo[2]{#2}
\providecommand\natexlab[1]{#1}
\providecommand\showeprint[2][]{arXiv:#2}

\bibitem[\protect\citeauthoryear{Alkouz, Bouguettaya, and Mistry}{Alkouz
  et~al\mbox{.}}{2020}]%
        {alkouz2020swarm}
\bibfield{author}{\bibinfo{person}{Balsam Alkouz}, \bibinfo{person}{Athman
  Bouguettaya}, {and} \bibinfo{person}{Sajib Mistry}.}
  \bibinfo{year}{2020}\natexlab{}.
\newblock \showarticletitle{Swarm-based drone-as-a-service (sdaas) for
  delivery}. In \bibinfo{booktitle}{\emph{IEEE ICWS}}.
  \bibinfo{pages}{441--448}.
\newblock


\bibitem[\protect\citeauthoryear{Bamburry}{Bamburry}{2015}]%
        {bamburry2015drones}
\bibfield{author}{\bibinfo{person}{Dane Bamburry}.}
  \bibinfo{year}{2015}\natexlab{}.
\newblock \showarticletitle{Drones: Designed for product delivery}.
\newblock \bibinfo{journal}{\emph{Design Management Review}}
  \bibinfo{volume}{26}, \bibinfo{number}{1} (\bibinfo{year}{2015}),
  \bibinfo{pages}{40--48}.
\newblock


\bibitem[\protect\citeauthoryear{Chmaj and Selvaraj}{Chmaj and
  Selvaraj}{2015}]%
        {chmaj2015distributed}
\bibfield{author}{\bibinfo{person}{Grzegorz Chmaj} {and} \bibinfo{person}{Henry
  Selvaraj}.} \bibinfo{year}{2015}\natexlab{}.
\newblock \showarticletitle{Distributed processing applications for UAV/drones:
  a survey}.
\newblock In \bibinfo{booktitle}{\emph{Progress in Systems Engineering}}.
  \bibinfo{publisher}{Springer}, \bibinfo{pages}{449--454}.
\newblock


\bibitem[\protect\citeauthoryear{Jones}{Jones}{2017}]%
        {jones2017international}
\bibfield{author}{\bibinfo{person}{Therese Jones}.}
  \bibinfo{year}{2017}\natexlab{}.
\newblock \bibinfo{booktitle}{\emph{International commercial drone regulation
  and drone delivery services}}.
\newblock \bibinfo{type}{{T}echnical {R}eport}. \bibinfo{institution}{RAND}.
\newblock


\bibitem[\protect\citeauthoryear{Kumar et~al\mbox{.}}{Kumar
  et~al\mbox{.}}{2021}]%
        {KUMAR20211}
\bibfield{author}{\bibinfo{person}{Adarsh Kumar} {et~al\mbox{.}}}
  \bibinfo{year}{2021}\natexlab{}.
\newblock \showarticletitle{A drone-based networked system and methods for
  combating coronavirus disease (COVID-19) pandemic}.
\newblock \bibinfo{journal}{\emph{FGCS}}  \bibinfo{volume}{115}
  (\bibinfo{year}{2021}), \bibinfo{pages}{1--19}.
\newblock


\bibitem[\protect\citeauthoryear{Shahzaad, Bouguettaya, Mistry, and
  Neiat}{Shahzaad et~al\mbox{.}}{2019}]%
        {shahzaad2019composing}
\bibfield{author}{\bibinfo{person}{Babar Shahzaad}, \bibinfo{person}{Athman
  Bouguettaya}, \bibinfo{person}{Sajib Mistry}, {and}
  \bibinfo{person}{Azadeh~Ghari Neiat}.} \bibinfo{year}{2019}\natexlab{}.
\newblock \showarticletitle{Composing drone-as-a-service (daas) for delivery}.
  In \bibinfo{booktitle}{\emph{IEEE ICWS}}. \bibinfo{pages}{28--32}.
\newblock


\bibitem[\protect\citeauthoryear{Shakhatreh et~al\mbox{.}}{Shakhatreh
  et~al\mbox{.}}{2019}]%
        {shakhatreh2019unmanned}
\bibfield{author}{\bibinfo{person}{Hazim Shakhatreh} {et~al\mbox{.}}}
  \bibinfo{year}{2019}\natexlab{}.
\newblock \showarticletitle{Unmanned aerial vehicles (UAVs): A survey on civil
  applications and key research challenges}.
\newblock \bibinfo{journal}{\emph{IEEE Access}}  \bibinfo{volume}{7}
  (\bibinfo{year}{2019}), \bibinfo{pages}{48572--48634}.
\newblock


\end{thebibliography}

\end{document}